\pdfoutput=1

\documentclass[11pt]{article}

\usepackage[final]{acl}

\usepackage{times}
\usepackage{latexsym}

\usepackage[T1]{fontenc}

\usepackage[utf8]{inputenc}

\usepackage{microtype}

\usepackage{inconsolata}

\usepackage[utf8]{inputenc} 
\usepackage[T1]{fontenc}    
\usepackage{hyperref}       
\usepackage{url}            
\usepackage{booktabs}       
\usepackage{amsfonts}       
\usepackage{nicefrac}       
\usepackage{microtype}      
\usepackage{xcolor}         

\usepackage{multirow}
\usepackage{colortbl}
\usepackage{amsmath}
\usepackage{amssymb}
\usepackage{pifont}
\usepackage{graphicx}
\usepackage{subcaption}

\usepackage{tcolorbox}
\tcbuselibrary{skins}

\definecolor{mygreen}{rgb}{0.0,0.6,0.0} 
\newcommand{\positive}[1]{\textcolor{red}{$\uparrow$#1}}
\newcommand{\negative}[1]{\textcolor{mygreen}{$\downarrow$#1}}
\newcommand{\xmark}{\ding{55}}
\title{A Scalable Multi-LLM Collaboration System with Retrieval-based Selection and Exploration-Exploitation-Driven Enhancement}

\author{
 \textbf{Shengji Tang\thanks{Equal contribution.}\thanks{Work done during the author’s internship at Shanghai Artificial Intelligence Laboratory.}\textsuperscript{1,2}},
 \textbf{Jianjian Cao\textsuperscript{$\ast$}\textsuperscript{$\dagger$}\textsuperscript{1,3}},
 \textbf{Weihao Lin\textsuperscript{3}},
 \textbf{Jiale Hong\textsuperscript{4}},
\\
 \textbf{Bo Zhang\textsuperscript{1}},
 \textbf{Shuyue Hu\textsuperscript{1}},
 \textbf{Lei Bai\textsuperscript{1}},
 \textbf{Tao Chen \textsuperscript{3}},
\\
 \textbf{Wanli Ouyang\textsuperscript{1,2}},
 \textbf{Peng Ye\textsuperscript{1,2}\thanks{Corresponding author}}
\\
 \textsuperscript{1}Shanghai Artificial Intelligence Laboratory,
 \textsuperscript{2}The Chinese University of Hong Kong,
 \\
 \textsuperscript{3}Fudan University,
 \textsuperscript{4}Shanghai Jiao Tong University
\\
 \small{
   \textbf{Correspondence:} \href{yepeng@pjlab.org.cn}{yepeng@pjlab.org.cn}
 }
}

\begin{document}
\maketitle
\begin{abstract}
\label{sec:ab}
Existing multi-LLM collaboration systems often encounter scalability challenges when integrating new LLMs and tasks, leading to suboptimal performance.
To address this, we propose SMCS, a Scalable Multi-LLM Collaboration System designed to effectively coordinate multiple open-source LLMs. 
The system consists of two core components: a Retrieval-based Prior Selection (RPS) module, which dynamically selects the most suitable LLMs for each input, and an Exploration–Exploitation-Driven Posterior Enhancement (EPE) module, which fosters response diversity and selects high-quality outputs through a hybrid scoring mechanism.
Experiments on eight mainstream benchmarks validate the effectiveness of our system: by integrating fifteen open-source LLMs, SMCS outperforms prevailing closed-source LLMs, e.g.,  \textit{GPT-4.1}(\textbf{+5.36\%}) and \textit{GPT-o3-mini}(\textbf{+5.28\%}) across multiple tasks. 
Remarkably, it even exceeds the average of best results on different datasets with open-source LLMs (\textbf{+2.86\%}), significantly advancing the empirical performance frontier of open-source collaboration. The code is released at \url{https://github.com/magent4aci/SMCS}. 
\end{abstract}
\section{Introduction}
\label{intro}
\begin{figure*}[t]
\centering
    \begin{subfigure}{0.45\linewidth}
      \centering
      \includegraphics[width=1.0\linewidth]{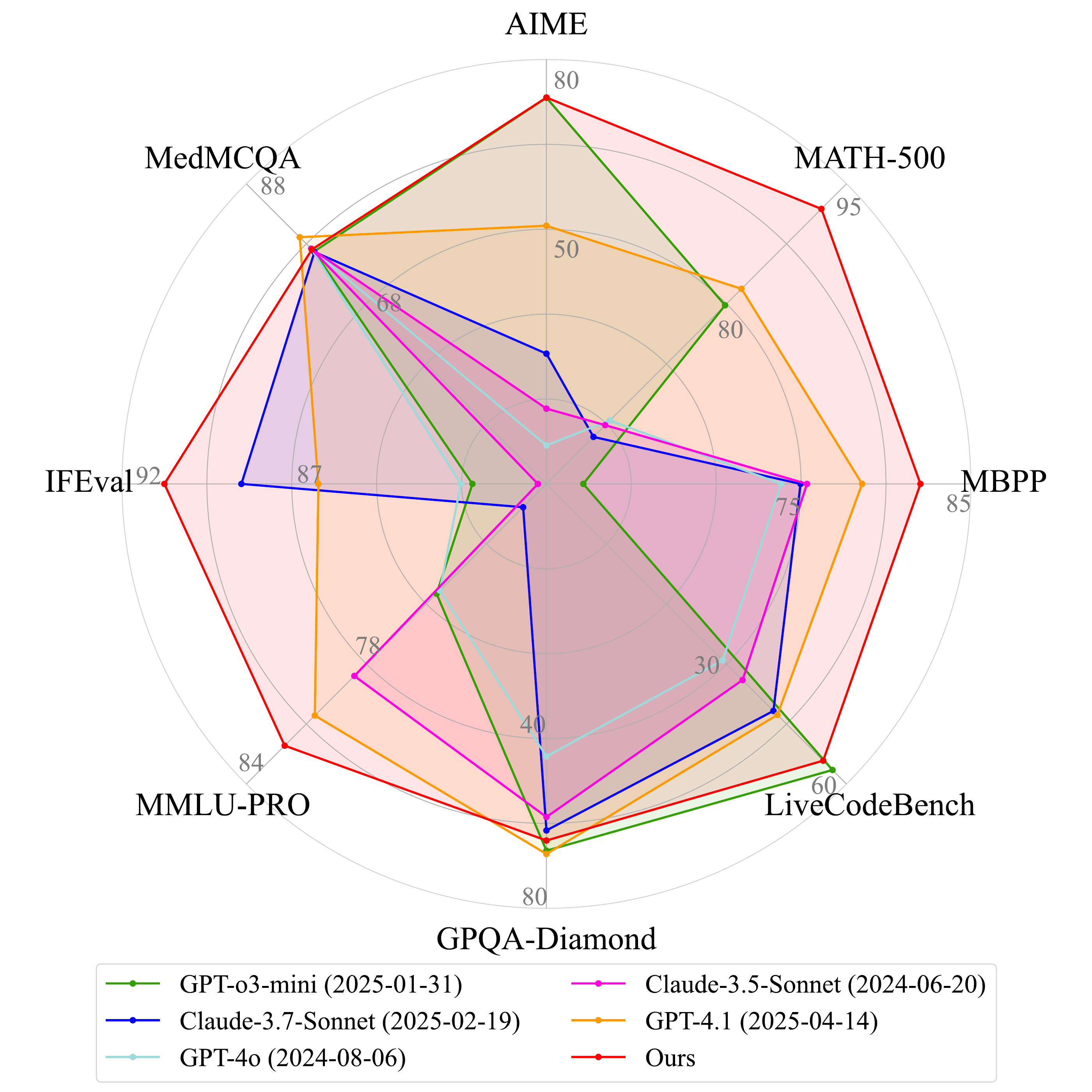}
      \caption{Comparisons with closed-source LLMs.}
      \label{fig:radar_close}
    \end{subfigure}
    \hspace{0.1\linewidth}
    \begin{subfigure}{0.4\linewidth}
      \centering
      \includegraphics[width=1.0\linewidth]{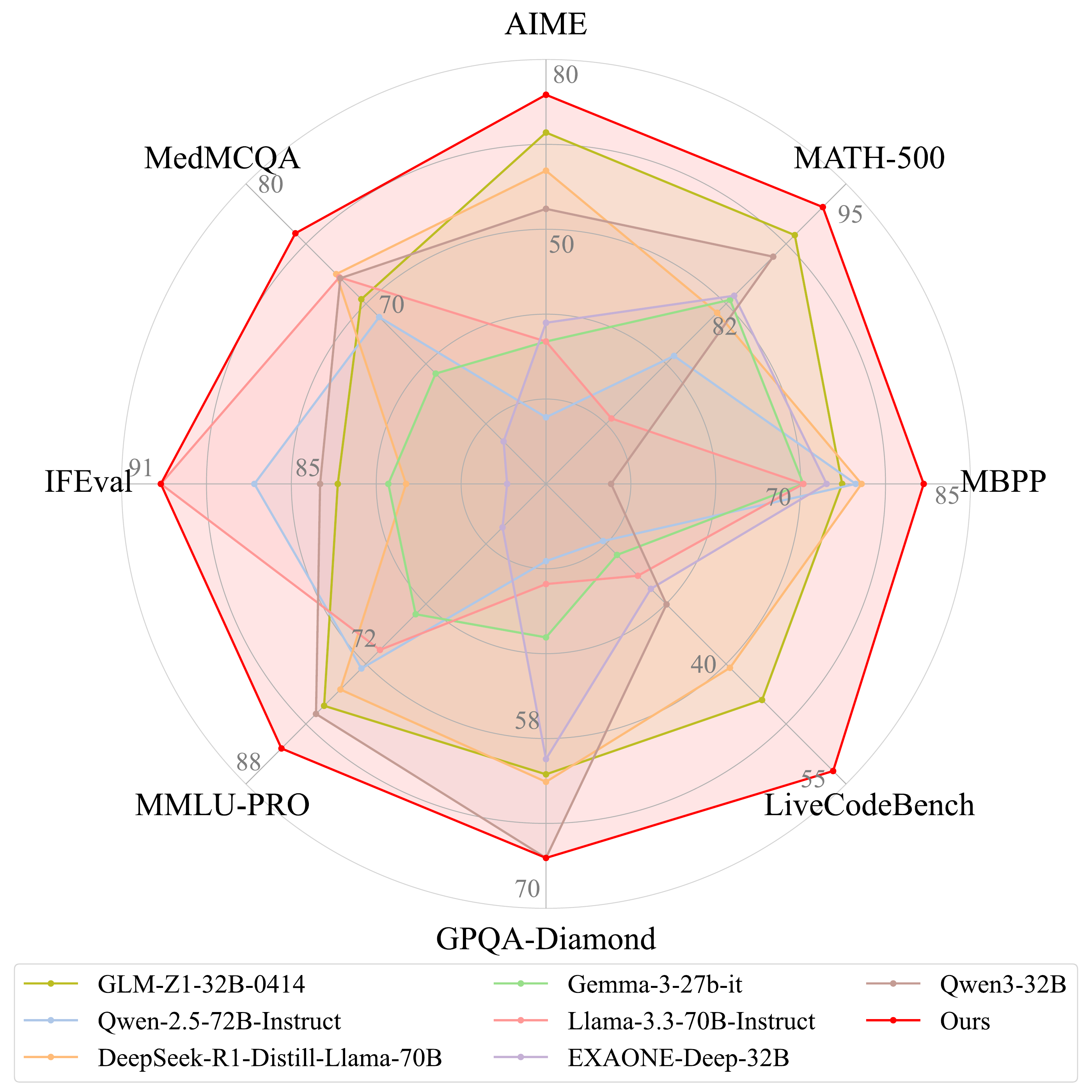}
      \caption{Comparisons with open-source LLMs.}
      \label{fig:radar_open}
    \end{subfigure}
    \vspace{-2mm}
\caption{Results on eight mainstream benchmarks. The proposed SMCS orchestrates fifteen open-source LLMs, surpassing both open-source and closed-source LLMs and pushing the upper bound of a single LLM.}
\label{fig:radar_cmp}
\end{figure*}

Recently, Large Language Models (LLMs)~\cite{gpt-4.1,Claude-3.5-sonnet,Claude-3.7-sonnet} have achieved remarkable success across diverse NLP tasks.
With the development of LLM training techniques, a growing number of heterogeneous LLMs, particularly open‑source LLMs trained on disparate data, have emerged. 
Due to structural diversity and bias in the training data, these LLMs possess diverse specialized skills and are expert in distinct areas. 
Therefore, a pivotal and valuable question naturally arises: how can we sustainably harness and scale up the vast and diverse collaboration of LLMs to continually push the performance frontier and advance collective intelligence?

\par To answer this question, a general approach is to construct a Multi-LLM Collaboration System (MCS). The MCS aims to orchestrate interactions among multiple LLMs, enable information exchange and integration, and generate high-quality responses. 
Emerging works have explored the construction of MCS, which can be broadly divided into two categories: (1) MCS via prior LLM selection. 
These approaches~\cite{chen2025symbolic,lu2023routing,shnitzer2023large,chen2024routerdc} select appropriate LLMs before response generation by leveraging prior knowledge corresponding to LLMs, such as their performance on standard benchmarks or model embeddings obtained from training on specific datasets. 
By selecting the most suitable models for each given question in advance, these methods aim to increase the likelihood of generating high-quality responses. 
(2) MCS via posterior response enhancement. These approaches~\cite{chen2024more,chen2023frugalgpt,gui2024bonbon,choudhury2025process} assess the quality of responses after each LLM has generated its answer, using inter- or intra-response criteria such as reward model scores, perplexity, or majority voting.
Due to performing reasoning, these methods provide a more accurate evaluation of response quality compared to relying solely on prior information. 

\par However, both categories of methods encounter challenges when scaling the number of LLMs and tasks.
For MCS based on prior LLM selection, they either require end-to-end router training~\cite{chen2024routerdc} for each individual LLM, making it difficult to continuously incorporate new LLMs, or rely on limited and discrete capability labels~\cite{chen2025symbolic}, which are insufficient for comprehensive analysis on a given question and hard to handle unseen questions. 
For MCS based on posterior response enhancement, these methods typically rely on a single posterior criterion, which can introduce bias and lead to inaccurate quality assessments. Moreover, they mainly focus on selecting from an existing pool of responses, lacking the ability to generate new and diverse high-quality responses, limiting their overall collective performance. Besides the above limitations, current MCS methods often fail to effectively integrate prior and posterior methods in a coupled manner, which causes unfiltered low-quality responses as bottlenecks, which significantly hinders the overall performance and scalability of the collaboration system.

\par To enhance the scalability and further advance the performance of MCS, we propose a novel framework called \underline{S}calable \underline{M}ulti-LLM \underline{C}ollaboration \underline{S}ystem (SMCS).
Specifically, we first construct a question bank comprising diverse questions from multiple domains, along with an LLM pool containing plentiful heterogeneous LLMs. 
Each LLM in the pool is evaluated on the question bank to record its response, representing its capacity across diverse domains. 
Further, inspired by Retrieval-Augmented Generation (RAG)~\cite{lewis2020retrieval,chen2024benchmarking}, we design a retrieval-based prior selection (RPS) strategy: given any question, we retrieve similar questions from the question bank. A weighted score is computed for each LLM based on its performance on the retrieved questions, which serves as the prior information for selecting high-scoring LLMs.
After that, we introduce exploration–exploitation-driven posterior enhancement (EPE): in the exploration phase, these responses are dropped via prior scores to form multiple answer subsets, which are independently aggregated by the selected LLM aggregator; in the exploitation phase, the aggregating responses are evaluated using a hybrid posterior scores of mean pairwise similarity and perplexity. The aggregated response with the highest score is selected as the final response.

We conduct extensive experiments to validate the effectiveness of the proposed framework across eight datasets. Notably, by jointly leveraging fifteen mid-sized open-source LLMs, SMCS significantly surpasses the current flagship closed-source models, such as GPT-4.1(\textbf{+5.36\%}) and GPT-o3-mini(\textbf{+5.28\%}).  Moreover, SMCS also exceeds both the average performance of the open-source best baselines(\textbf{+2.86\%}). This demonstrates the strong capability of SMCS and its potential to break through the upper bound of performance. Besides, SMCS can consistently obtain gains without remarkable saturation by progressively increasing the number of LLMs, demonstrating excellent scalability.
Our contributions are summarized as follows: 
\vspace{-2mm}
\begin{itemize}
\item We first present a comprehensive analysis of existing multi-LLM collaboration systems from prior and posterior perspectives, and identify several key limitations hindering the development of scalable and high-performance MCS frameworks.
\vspace{-2mm}
\item We propose SMCS, a scalable multi-LLM collaboration framework. It jointly considers prior and posterior information, where a retrieval-based prior selection strategy is proposed to recruit suitable LLMs at the instance level, and an exploration–exploitation-driven posterior enhancement strategy is designed to generate higher-quality responses.
\vspace{-2mm}
\item Extensive experiments across diverse datasets validate the scalability and effectiveness of SMCS, demonstrating its ability to enable continuous expansion of LLMs while harnessing open-source models to surpass prevailing closed-source models.
\end{itemize}
\vspace{-1mm}

\begin{figure*}[t]
  \centering
  \includegraphics[width=\linewidth]{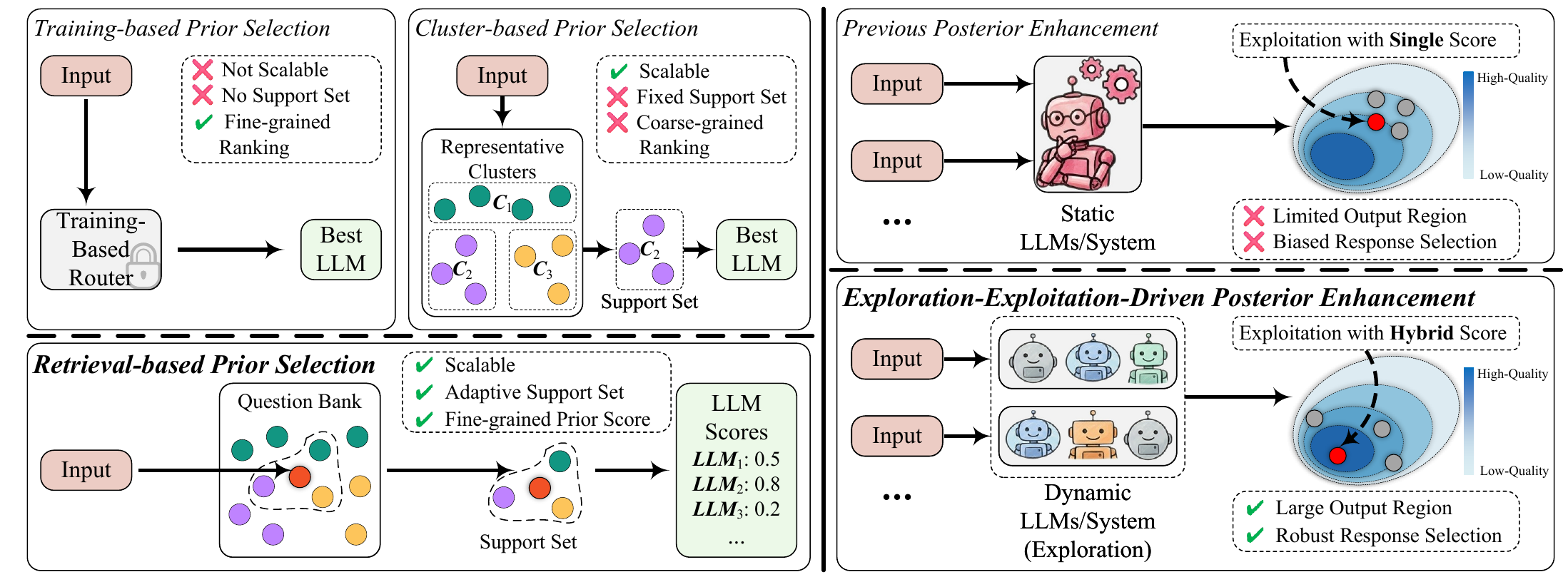}
  \vspace{-7mm}
  \caption{The illustration of two core innovations in proposed SMCS. SMCS adopts different and more advanced paradigms for prior selection and posterior enhancement, achieving significant scalability and performance.}
  \label{fig:simple_comparison}
\end{figure*}

\section{Related works}
\label{relatedwork}

\textbf{Prior-based LLM Collaboration.}
Prior-based methods
focus on dynamically selecting or routing LLMs before generating responses. 
%
%
Recent research explores LLM routing, where a selector determines the most suitable model for a given question without integrating all LLMs. The preliminary work~\cite{shnitzer2023large} proposes binary classifiers to predict the correctness of individual LLMs, while ZOOTER~\cite{lu2023routing} aligns a router with reward-model supervision. RouterDC~\cite{chen2024routerdc} utilizes dual contrastive learning for improved accuracy. While GraphRouter~\cite{feng2025graphroutergraphbasedrouterllm} constructs model selection as a dynamic link prediction problem by constructing heterogeneous task-query-LLM graphs with GNNs, MODEL-SAT~\cite{zhang2025capability} focuses on performance-based capability representations. The latter specifically leverages a lightweight LLM to predict the most effective candidate for a given task.
Most relevant to our work, Symbolic\_MoE~\cite{chen2025symbolic} proposes a Mixture-of-Experts framework that dynamically selects and combines LLMs 
based on skill-specific expertise.

\textbf{Posterior-based LLM Collaboration.}
Posterior based methods
aggregate outputs from multiple LLM executions to derive an improved response. Simple but effective techniques such as Voting~\cite{li2024more,wang2022self} and advanced ranking-based approaches such as LLM-Blender~\cite{jiang2023llm}, demonstrate the benefits of ensemble refinement. 
%
Besides, techniques like majority voting~\cite{chen2024more}, self-consistency~\cite{wang2022self,chen2023universal}, and best-of-n sampling~\cite{gui2024bonbon} could enhance reliability in tasks lacking verification tools. 
Mixture of Agents (MoA)~\cite{wang2024mixture} introduces a framework for combining LLM agents into ensembles, relying on a fixed set of agents across tasks. 
Similarly, Self-MoA~\cite{li2025rethinking} argues that invoking a single high-performing model multiple times, paired with an optimal aggregator, can achieve competitive performance without leveraging diverse LLMs.

While existing MCS demonstrate effectiveness, they suffer from two critical limitations: (1) scalability constraints that hinder seamless integration of new LLMs, and (2) suboptimal performance due to inefficient utilization and limited exploration of different LLMs' responses.
In this work, we propose SMCS that incorporates the advantages of prior and posterior approaches. It enables scalable instance-level LLM selection via RPS strategy, and extends the diversity of responses while making full use of them via designed EPE. 

\section{Method}
\label{method}
In this section, we first provide an overview of SMCS in Sec.~\ref{sec:overall_framework}. 
Then, the construction of the question bank is stated in Sec.~\ref{sec:question_bank}. Next, we present the retrieval-based prior selection (RPS) and exploration–exploitation-driven posterior enhancement (EPE) in Sec.~\ref{sec:prior_selection} and Sec.~\ref{sec:posterior_enhancement}. A visual comparison between proposed techniques and existing methods is shown in Fig.~\ref{fig:simple_comparison}.

\subsection{Overall Framework}
\label{sec:overall_framework}
\begin{figure*}[t]
  \centering
  \includegraphics[width=\linewidth]{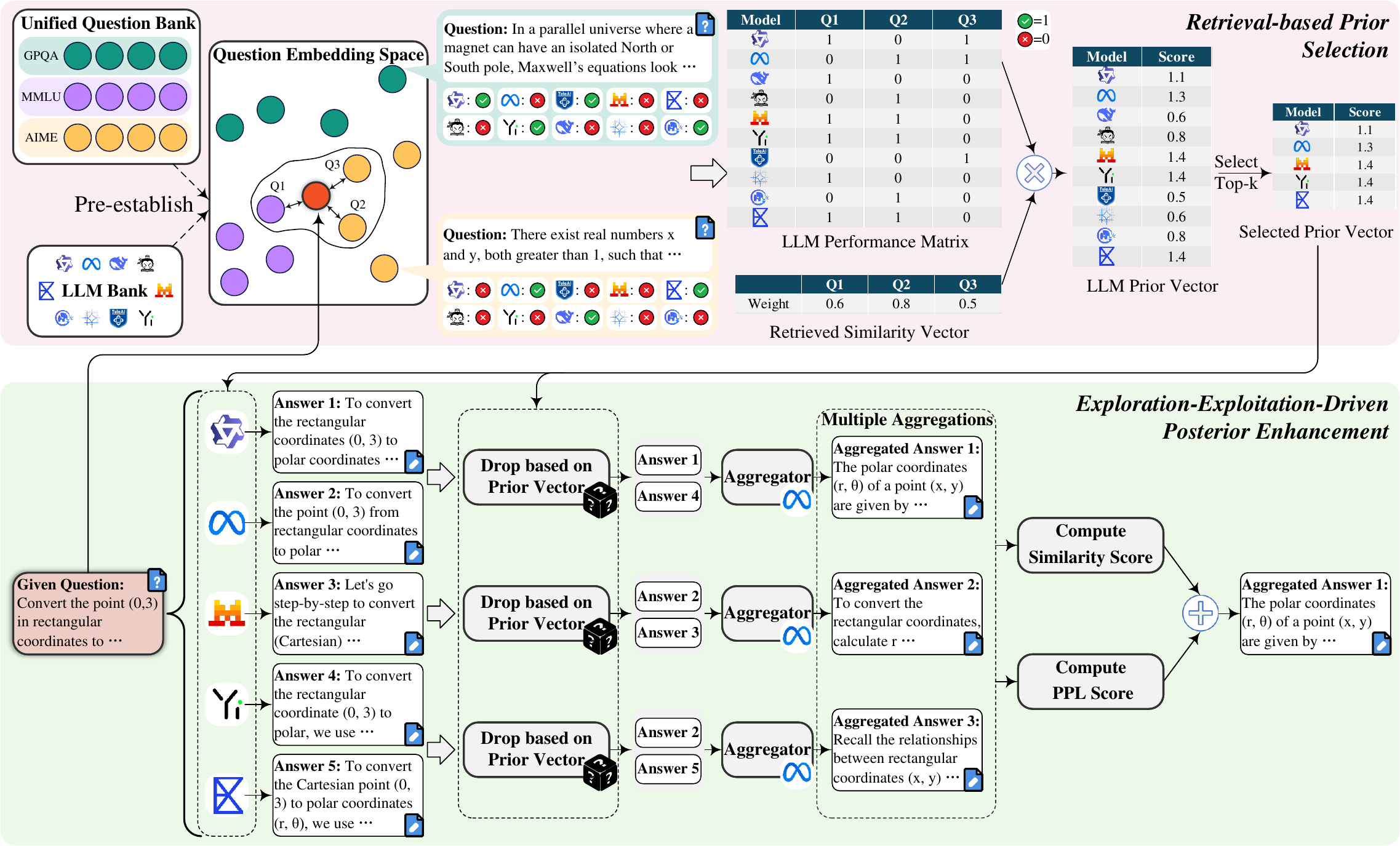}
  \vspace{-7mm}
  \caption{Overview of our SMCS framework. It dynamically selects Top-K expert LLMs from the predefined LLM bank through RPS module, then optimizes responses via EPE module to generate high-quality outputs.
  }
  \label{fig:framework}
\end{figure*}

As shown in Fig.~\ref{fig:framework}, for scalable and generalizable capability assessment for each LLM, SMCS constructs a unified question bank by integrating questions from multiple domains with their labels. Each LLM is evaluated on the unified question bank to obtain a fine-grained assessment of its capability distribution, which contains prior information of each LLM. 
During inference, SMCS consists of two stages: (1) Retrieval-based Prior Selection (2) Exploration-exploitation-driven Posterior Enhancement. In the first stage, given a question, SMCS retrieves related questions from the question bank to obtain prior information of each LLM and selects suitable expert LLMs as referencers. Then, all referencers are forwarded to collect their responses as references. Meanwhile, SMCS selects the LLM with the strongest instruction-following capability to serve as the aggregator. In the second stage, the references are dropped based on the distribution of the prior information of the corresponding referencers to generate multiple reference subsets. Each subset is aggregated by the aggregator, resulting in multiple candidates to explore high-quality responses. Finally, SMCS evaluates each candidate using a hybrid posterior score that incorporates both intra-response and inter-response criteria, serving as an exploitation over the output space of the aggregator. The candidate with the highest score is selected as the final response.

\subsection{Unified Question Bank}
\label{sec:question_bank}
Due to the heterogeneity of LLMs from various sources, it is infeasible to extract prior information by directly analyzing their architectures or parameters. To guarantee the generalization of prior information extraction across diverse LLMs and tasks, SMCS adopts a black-box evaluation strategy that analyzes the responses generated by each LLM to specific inputs. Specifically, given an LLM bank $\mathcal{A}=\{A_1, A_2, ...A_R\}$ containing $R$ LLMs, SMCS constructs a unified question bank $\mathcal{B}=\{(x_i^{qb}, y_i^{qb})|i \in [1, N]\}$ by sampling $N$ questions from the validation sets of diverse tasks for a comprehensive capability assessment for each LLM. $x_i^{qb}$ and $y_i^{qb}$ are the $i_{th}$ question and the corresponding label in the question bank, respectively. After constructing the unified question bank, each LLM $A_i$ is forwarded to answer all questions in the question bank, obtaining a capability vector $V_i^{qb} \in \{0,1\}^{N \times 1}$ that represents its capabilities across diverse tasks,
\begin{equation}
\resizebox{0.87\hsize}{!}{$
\begin{aligned}
V_i^{qb}=\left [\mathbf{1}_{\{A_i(x_0^{qb})=y_0^{qb}\}}, 
\mathbf{1}_{\{A_i(x_1^{qb})=y_1^{qb}\}},...,\mathbf{1}_{\{A_i(x_N^{qb})=y_N^{qb}\}}\right ]^\top
\end{aligned}
$}
\end{equation}
where $\mathbf{1}_{\{\cdot\}}$ is the indicator. It is worth noting that for notational simplicity, the parameters $\theta_i$ of $A_i$ are omitted, and we use ``$=$'' to represent verifying the correctness of a response. Moreover, a pre-trained embedding model $\mathcal{M}_{emb}$ is introduced to embed each question $x_i^{qb}$ into latent space for the later retrieval, denoted as $e_i^{qb} = Norm(\mathcal{M}_{emb}(x_i^{qb})) \in \mathbb{R}^{d \times 1}$, where $d$ is the embedding dimension of $\mathcal{M}_{emb}$ and $Norm(\cdot)$ is normalization function. The capability vector $V_i^{qb}$ records the historical performance of each LLM at the instance level, providing fine-grained prior information. 

\subsection{Retrieval-based Prior Selection}
\label{sec:prior_selection}
The key to selecting the optimal LLMs is establishing the relevance between the given question and the collected prior information. Existing methods typically introduce a preprocessing procedure and assign the given question to an explicit or implicit category based on unsupervised clustering~\cite{jitkrittum2025universal,srivatsa2024harnessing} or supervised learning~\cite{shnitzer2023large,chen2024routerdc}.
The prior information associated with that category is used to estimate the capabilities of different LLMs for the given question. However, the complex preprocessing introduces noise and bias, potentially incorporating irrelevant prior information. To address these issues, inspired by the Retrieval-Augmented Generation (RAG)~\cite{lewis2020retrieval,chen2024benchmarking} paradigm, we design a retrieval-based prior selection without complex preprocessing. The core idea is to retrieve questions similar to the given question as support questions and then utilize the weighted scores on the support questions as a prior representation of LLMs' capabilities. Specifically, given a question $x^{in}$, a embedding model $\mathcal{M}_{emb}$ transfer it into an embedding vector $e^{in}=Norm(\mathcal{M}_{emb}(x^{in})) \in \mathbb{R}^{d \times 1}$. Then, a cosine similarity of $e^{in}$ with all $e^{qb}$ is computed to obtain similarity vector $S^{in} \in [0,1]^{N \times 1}$, denoted as 
\begin{align}
    S^{in} = [e^{qb}_{1},e^{qb}_{2},...,e_N^{qb}]^\top e^{in} .
\label{formu:sim}
\end{align}
To adaptively retrieve the support questions, a base number $N^{sup\_base}$ is defined to ensure sufficient evaluation coverage. Moreover, a tolerance threshold coefficient $\gamma \in [0,1]$ is introduced to obtain a relative threshold to select the support questions. The index of support questions is denoted as 
\begin{align}
    I^{sup} = \{i|S^{in}[i] \ge \gamma \mathrm{max}_{N^{sup\_base}}(S^{in})\}
\end{align}
where $\mathrm{max}_{k}(\cdot)$ refers to the $k_{th}$ largest element in a vector. The number of support questions is $N^{sup}=|I|$. Then, according to $I^{sup}$, the retrieved similarity vector ${\hat{S}^{in}} \in [0,1]^{N^{sup} \times 1}$ can be indexed from $S^{in}$, denoted as $\hat{S}^{in}=S^{in}_I$, and LLM performance matrix $M^{qb} \in \{0,1\}^{R \times N_{sup}}$ can be indexed from LLM capability vector $V^{qb}$, denoted as $M^{qb}=[V^{qb}_{I,1}, V^{qb}_{I,2}, ..., V^{qb}_{I,R}]^\top$, where $R$ is the number of LLMs in LLM bank. The LLM prior vector $V^{ref} \in \mathbb{R}^{R \ \times 1}$ can be computed by 
\begin{align}
    V^{ref} = M^{qb}\hat{S}^{in}.
    \label{formu:prior_vector}
\end{align}
Given the number of selected referencers $K$, the selected prior vector can be denoted as
\begin{align}
    \hat{V}^{ref}=V^{ref}_{I^{ref}}, I^{ref}=\mathrm{argtop}_{K}(V^{ref}),
\end{align}
where $\mathrm{argtop}_K(\cdot)$ refers to obtaining the indices of the largest K elements of a vector. The LLMs with the indices $I^{ref}$ are selected as referencers in the inference, denoted as $\mathcal{A}^{ref} = \{ A_i|i \in I^{ref}\}$.

\begin{table*}[!t]
  \centering
  \vspace{-3mm}
  \resizebox{\textwidth}{!}{
    \begin{tabular}{lccccccccc}
      \toprule
      \textbf{Model} & \textbf{AIME} & \textbf{MATH-500} & \textbf{MBPP} & \textbf{LiveCodeBench} & \textbf{GPQA-Diamond} & \textbf{MMLU-PRO} & \textbf{IFEval} & \textbf{MedMCQA} & \textbf{Avg} \\
      \midrule
      \rowcolor{gray!20} \multicolumn{10}{c}{\textit{Close-source LLMs}} \\
      GPT-o3-mini(2025-01-31)~\cite{gpt-o3-mini} & 73.33 & 84.40 & 62.00 & 54.70 & 66.67 & 74.00 & 82.00 & 74.92 & 71.50 \\
      Claude-3.7-Sonnet(2025-02-19)~\cite{Claude-3.7-sonnet} & 26.70 & 73.20 & 75.40 & 41.30 & 63.64 & 69.43 & 88.00 & 74.75 & 64.05 \\
      GPT-4o(2024-08-06)~\cite{achiam2023gpt} & 10.00 & 74.60 & 74.20 & 29.80 & 52.53 & 73.83 & 82.30 & 76.17 & 59.18 \\
      Claude-3.5-Sonnet(2024-06-20)~\cite{Claude-3.5-sonnet} & 16.70 & 74.20 & 75.80 & 34.30 & 61.62 & 78.34 & 80.30 & 76.00 & 62.16 \\
      GPT-4.1(2025-04-14)~\cite{gpt-4.1} & 50.00 & 85.80 & 79.20 & 42.20 & 67.17 & 80.43 & 86.00 & 80.58 & 71.42 \\
      Close-source Average & 35.34 & 78.44 & 73.32 & 40.46 & 62.33 & 75.21 & 83.72 & 76.48 & 65.66 \\
      \rowcolor{gray!20} \multicolumn{10}{c}{\textit{Open-source LLMs}} \\
      GLM-Z1-32B-0414~\cite{glm2024chatglm} & 66.70 & 90.00 & 74.40 & 44.40 & 59.60 & 76.76 & 83.00 & 70.50 & 70.67 \\
      Qwen-2.5-72B-Instruct~\cite{qwen2.5} & 16.70 & 78.80 & 75.80 & 26.10 & 45.45 & 72.16 & 86.30 & 69.08 & 58.80 \\
      DeepSeek-R1-Distill-Llama-70B~\cite{deepseekai2025} & 60.00 & 82.80 & 76.40 & 40.70 & 60.10 & 74.75 & 80.30 & 72.50 & 68.44 \\
      QwQ-32B~\cite{qwq-32b-preview} & 46.70 & 87.80 & 81.80 & 38.60 & 57.07 & 74.67 & 81.70 & 69.83 & 67.27 \\
      Gemma-3-27b-it~\cite{team2024gemma} & 30.00 & 84.00 & 70.40 & 27.70 & 50.51 & 65.47 & 81.00 & 64.58 & 59.21 \\
      Qwen2.5-32b-Instruct~\cite{Qwen2.5-32B-Instruct} & 20.00 & 75.60 & 76.00 & 24.00 & 40.91 & 69.15 & 78.70 & 62.92 & 55.91 \\
      TeleChat2-35B-32K~\cite{wang2024telechat} & 10.00 & 70.00 & 70.00 & 19.50 & 33.33 & 67.98 & 82.00 & 57.08 & 51.24 \\
      InternLM2.5-20B-Chat~\cite{cai2024internlm2} & 3.30 & 55.20 & 55.00 & 14.90 & 34.85 & 44.23 & 64.70 & 51.92 & 40.51 \\
      Llama-3.3-70B-Instruct~\cite{grattafiori2024llama} & 30.00 & 73.00 & 70.40 & 30.10 & 46.97 & 69.87 & 90.00 & 72.25 & 60.32 \\
      EXAONE-Deep-32B~\cite{exaone-deep} & 33.30 & 84.38 & 72.80 & 31.60 & 58.59 & 54.76 & 76.30 & 59.17 & 58.86 \\
      Qwen2.5-Coder-32B-Instruct~\cite{hui2024qwen2} & 16.70 & 73.60 & 78.00 & 27.70 & 41.92 & 61.79 & 80.30 & 57.25 & 54.66 \\
      Qwen3-32B~\cite{qwq32b} & 53.30 & 88.00 & 50.60 & 33.40 & 65.15 & 77.76 & 83.70 & 72.17 & 65.51 \\
      Llama-3.3-Nemotron-Super-49B-v1~\cite{grattafiori2024llama} & 16.70 & 75.20 & 65.40 & 28.00 & 48.48 & 67.47 & 82.70 & 70.92 & 56.86 \\
      DeepSeek-R1-Distill-Qwen-32B~\cite{deepseekai2025} & 56.70 & 85.60 & 81.00 & 44.70 & 60.10 & 75.17 & 73.70 & 67.25 & 68.03 \\
      HuatuoGPT-o1-72B~\cite{chen2024huatuogpto1medicalcomplexreasoning} & 16.70 & 73.00 & 78.00 & 27.40 & 50.00 & 74.16 & 74.00 & 75.25 & 58.56 \\
      Open-source Average & 31.79 & 78.47 & 71.73 & 30.59 & 50.20 & 68.41 & 79.89 & 66.18 & 59.66 \\
      \rowcolor{gray!20} \multicolumn{10}{c}{\textit{Other Methods}} \\
      Symbolic-MoE*~\cite{chen2025symbolic} &  50.00   &  90.40  &  82.60   &  43.01  &  62.63    &           80.60  & 89.00  &  74.88 & 71.64\\
      MoA~\cite{wang2024mixture} &  53.33   &    87.80       &  82.00   &    40.12           &    58.80         &   79.6        &    89.33    &   73.08      &  70.51  \\
      Self-MoA~\cite{li2025rethinking} &  76.67   &  93.00  &  83.20   &  29.39  &   64.41  &  69.89   &  86.00 & 74.88  &  72.18  \\
      Self Consistency (Best on Validation) ~\cite{chen2023universal} &  60.00  &   90.40   &   82.40   &   40.12  & 63.64  & 78.01   &  90.39   &  74.83  &  72.47 \\
      Majority Voting~\cite{chen2024more} &  56.67   &     90.2      &   80.4  &     34.65          &    26.26         &  80.85         &   80.67     &    73.33     &  65.38  \\
      Simple Router & 46.70    &     88.00      &  81.80   &      33.40         &    60.10        &     72.50      &    90.00    &   72.50      &  68.40  \\
      \rowcolor{gray!20} \multicolumn{10}{c}{\textit{Ours v.s. Strong Baselines}} \\
      Open-source Upper Bound & 66.70 & 90.00 & 81.80 & 44.70 & 65.15 & 77.76 & 90.00 & 75.25 & 73.92 \\
      \textbf{SMCS(ours)} & \textbf{73.33} & \textbf{92.60} & \textbf{82.80} & 52.58 & 65.15 & \textbf{82.02} & \textbf{90.00} & 75.75 & \textbf{76.78} \\
      \textit{- v.s. Self Consistency (Best on Validation)} & \positive{13.33} & \positive{2.20} & \positive{0.40} & \positive{12.46} & \positive{1.51} & \positive{4.01} & \negative{0.39} & \positive{0.92} & \positive{4.31} \\
      \textit{- v.s. MoA} & \positive{20.00} & \positive{4.80} & \positive{0.8} & \positive{12.46} & \positive{6.35} & \positive{2.42} & \positive{0.67} & \positive{2.67} & \positive{6.27} \\
      \textit{- v.s. GPT-o3-mini} & 0 & \positive{8.20} & \positive{20.80} & \negative{2.12} & \negative{1.52} & \positive{8.02} & \positive{8.00} & \positive{0.83} & \positive{5.28} \\
      \bottomrule
    \end{tabular}
  }
    \caption{Main Results of our SMCS framework with fifteen open-source LLMs on eight mainstream benchmarks.}  
  \label{tab:main_results}  
\end{table*}

\subsection{Exploration-Exploitation-Driven Posterior Enhancement}
\label{sec:posterior_enhancement}
After prior selection, SMCS is required to further evaluate and organize references to filter out inferior information and generate higher-quality responses. 
Due to differences in training data and architectures, reference responses differ significantly in patterns and distributions, making direct posterior evaluation challenging. 
%
To address these issues, we adopt an exploration-exploitation-driven posterior enhancement strategy. It explore diverse and high-quality aggregations by dropping some inferior references and aggregating multi times based on prior information, and exploits the aggregations by introducing a hybrid posterior score to select the optimal aggregation as final response. Specifically, given the referencers LLMs $\mathcal{A}^{ref}$ from prior selection, the references can be collected by forwarding all referencers, denoted as $O^{all}=\{A_i(x^{in})|A_i \in \mathcal{A}^{ref}\}$. For exploration, given a dropping number $K_{drop}$, the references are dropped following the prior-based discrete sampling distribution $\mathcal{D}$, which is denoted as 
\begin{equation}
\resizebox{0.8\hsize}{!}{$
\begin{aligned}
    \mathcal{D}=\left [ \frac{e^{\widetilde{\hat{V}^{ref}}[1]}}{\sum_{j=1}^{K} e^{\widetilde{\hat{V}^{ref}}[j]}}, \frac{e^{\widetilde{\hat{V}^{ref}}[2]}}{\sum_{j=1}^{K} e^{\widetilde{\hat{V}^{ref}}[j]}},...,
    \frac{e^{\widetilde{\hat{V}^{ref}}[K]}}{\sum_{j=1}^{K} e^{\widetilde{\hat{V}^{ref}}[j]}} \right ], \\
    \widetilde{\hat{V}^{ref}}=\left [\frac{\hat{V}^{ref}[1]-\overline{\hat{V}^{ref}}}{std(\hat{V}^{ref})}, \frac{\hat{V}^{ref}[2]-\overline{\hat{V}^{ref}}}{std(\hat{V}^{ref})},...,\frac{\hat{V}^{ref}[K]-\overline{\hat{V}^{ref}}}{std(\hat{V}^{ref})} \right ]
\end{aligned}
$}
\end{equation}
where $std(\cdot)$ refers to obtaining the standard deviation, and we use a renormalize-after-each-draw rule~\cite{panahbehagh2021sequential} to achieve successive unequal-probability sampling~\cite{yu2012inclusion}, which can be seen as sampling $K-K_{drop}$ references from $O^{all}$ following $\mathcal{D}$ without replacement. After prior dropping $n$ times, multiple subsets $O^{sub}$ of $O^{all}$ are obtained, and each $O^{sub}$ is aggregated by an aggregator $A_{agg}$ to generate a aggregation set, denoted as $G_i=A_{agg}(cat(O^{sub}_i))$ where $cat(\cdot)$ refers to concatenating the references and injecting prompts for aggregating. Then, the mean pairwise similarity of an aggregation $G_i$ is computed as a similarity score $\mathcal{S}^{sim}_{i}$, denoted as $\mathcal{S}^{sim}_{i}=\frac{1}{n}\sum_{j=1}^{n}sim(G_i,G_j)$, where $sim(\cdot,\cdot)$ is computing cosine similarity using embedding model same as Formulation~\ref{formu:sim}. Meanwhile, the perplexity score $\mathcal{S}^{PPL}_{i}$ is computed as $\mathcal{S}^{PPL}_{i}=1-PPL(G_i)$, where $PPL(\cdot)$ refers to computing the perplexity~\cite{parsing2009speech,hu2024can} of a response. Finally, the total score of an aggregation can be denoted as 
\begin{align}
    \mathcal{S}^{total} = \mathcal{S}^{sim} + \lambda \mathcal{S}^{PPL},
\end{align}
where $\lambda$ is the balance coefficient. Finally, the aggregation $G$ with the highest $\mathcal{S}^{total}$ is regarded as the final response of SMCS.

\section{Experiments}
\label{experiments}

\begin{figure*}[!t]
  \centering
  \includegraphics[width=0.95\linewidth]{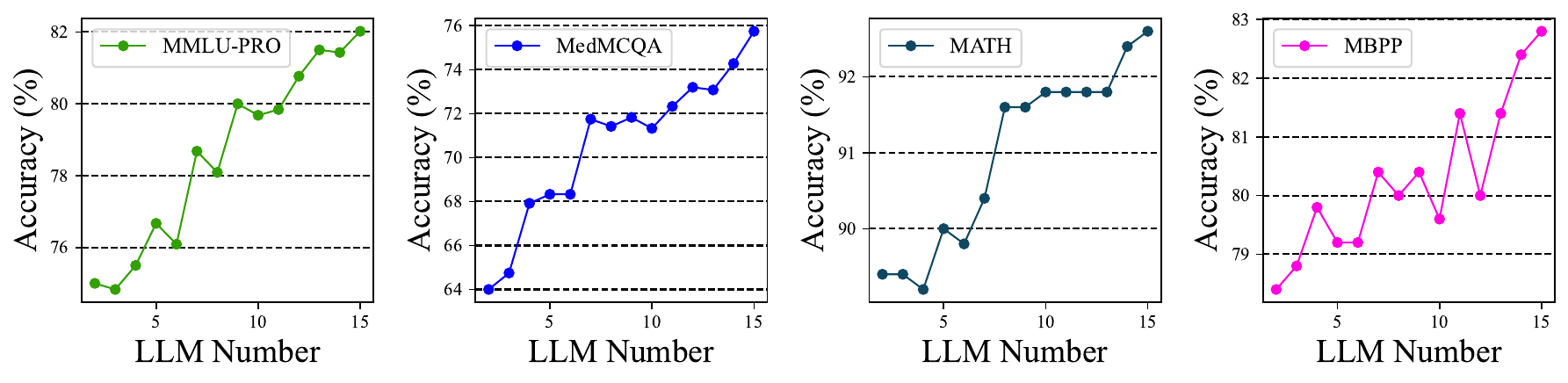}
  \vspace{-3mm}
  \caption{The scalability curve of SMCS. It can increasingly incorporate more LLMs for higher performance.}
  \label{fig:scale_ability}
\end{figure*}

\subsection{Experimental Setting}
\label{sec:exp_setting}
\noindent \textbf{Datasets.}
We establish a multi-domain evaluation comprising eight mainstream benchmarks spanning four key task categories: (1) Mathematical Problem Solving (MATH-500~\cite{hendrycks2021measuring}, AIME2024~\cite{AIME2024}), (2) Complex Reasoning (GPQA~\cite{rein2024gpqa}, MMLU-PRO~\cite{wang2024mmlu}, MedMCQA~\cite{pal2022medmcqa}), (3) Instruction Following (IFEval~\cite{zhou2023instruction}), and (4) Code Generation (MBPP~\cite{austin2021program}, LiveCodeBench~\cite{jain2024livecodebench}).
Each dataset is split into non-overlapping validation and test sets, with all validation sets combined to form the unified question bank for all benchmarks. 
See Appendix~\ref{app:data_details} for more details.

\noindent \textbf{LLM Bank.}
To achieve a balance between model diversity and efficiency, we carefully curate a collection of fifteen mid-sized open-source LLMs (from 20B to 72B) from various architectural families. 
SMCS framework employs a two-tiered model utilization strategy: reference models are dynamically selected from the full LLM bank during inference via task requirements, while the critical aggregator model is handled by Llama-3.3-70B-Instruct due to its exceptional instruction-following performance. See Appendix~\ref{app:llm_bank_details} for more details.

\vspace{-2mm}
\subsection{Main Results}
\label{mainresults}
As demonstrated in Table~\ref{tab:main_results}, our proposed SMCS framework establishes new state-of-the-art results across eight diverse benchmarks. 
Through comprehensive comparisons with (1) five leading close-source models (including GPT-o3-mini~\cite{gpt-o3-mini}, GPT-4.1~\cite{gpt-4.1}, GPT-4o~\cite{achiam2023gpt}, Claude-3.5-Sonnet~\cite{Claude-3.5-sonnet}, Claude-3.7-Sonnet~\cite{Claude-3.7-sonnet}), (2) fifteen representative open-source models, and (3) six existing collaboration methods, our approach demonstrates consistent and substantial improvements across all evaluation dimensions.
For example, our SMCS framework achieves 76.78\% average accuracy on eight benchmarks, representing substantial gains of +11.12\% and +17.12\% over the average closed-source (65.66\%) and open-source (59.66\%) baselines, respectively. 
Compared to existing collaboration approaches, SMCS outperforms Symbolic-MoE*~\cite{chen2025symbolic} by +5.14\%, MoA~\cite{wang2024mixture} by +6.27\%, Self-MoA~\cite{li2025rethinking} by +4.6\%. 
Remarkably, our solution even exceeds open-source upper bounds (+2.86\%), while significantly surpassing individual leading models, including GPT-4.1 (+5.36\%), GPT-4o (+17.60\%), and Claude-3.5-Sonnet (+12.73\%). 
It demonstrates that SMCS can effectively combine the strengths of multiple LLMs to achieve unprecedented performance.

\subsection{Efficiency Analysis}
Although SMCS focuses on exploring the maximum performance boundary of multi-LLM collaboration rather than optimizing efficiency, we further report the API cost and average query latency of multi-LLM methods and leading closed-source LLMs. As shown in Table~\ref{tab:efficiency}, SMCS achieves remarkable performance superiority, e.g., +5.14\% and +5.28\% compared with Symbolic-MoE and GPT-o3-mini, with a competitive cost and inference time. It verifies the feasibility and economy of SMCS in practical implementation. The efficiency of SMCS comes from 1) APIs of mid-sized open-source LLMs are dramatically cheaper than closed-source LLMs; 2) Although SMCS requires more LLM forward passes, most of these forward passes, e.g., the inferences of different referencers and aggregating multiple times, are independent and can be parallelized, making the overall inference time only determined by the slowest LLM.  

\begin{table}[]
\resizebox{0.46\textwidth}{!}{
\begin{tabular}{l|ccc}
\hline
Method/Model                          & Avg Acc(\%) & Cost(\$) & Avg Latency(s) \\ \hline
GPT-o3-mini                           & 71.50        & 15.36    & 10.42              \\
Claude-3.7-Sonnet                     & 64.05       & 20.38    & 17.92              \\
GPT-4.1                               & 71.42       & 11.98    & \textbf{10.11}               \\
MoA                                   & 70.51       & 9.42     & 18.64              \\
Self MoA                              & 72.18       & 9.14     & 12.82              \\
Symbolic-MoE                          & 71.64       & \textbf{7.86}     & 12.47              \\
Self Consistency(Best on Validation) & 72.47       & 8.39     & 12.78              \\ \hline
SMCS(ours)                            & \textbf{76.78}       & 8.11     & 12.32               \\ \hline
\end{tabular}
}
\caption{Cost and average latency of different methods.}
\label{tab:efficiency}
\end{table}

\begin{figure*}[!t]
  \centering
  \includegraphics[width=0.9\linewidth]{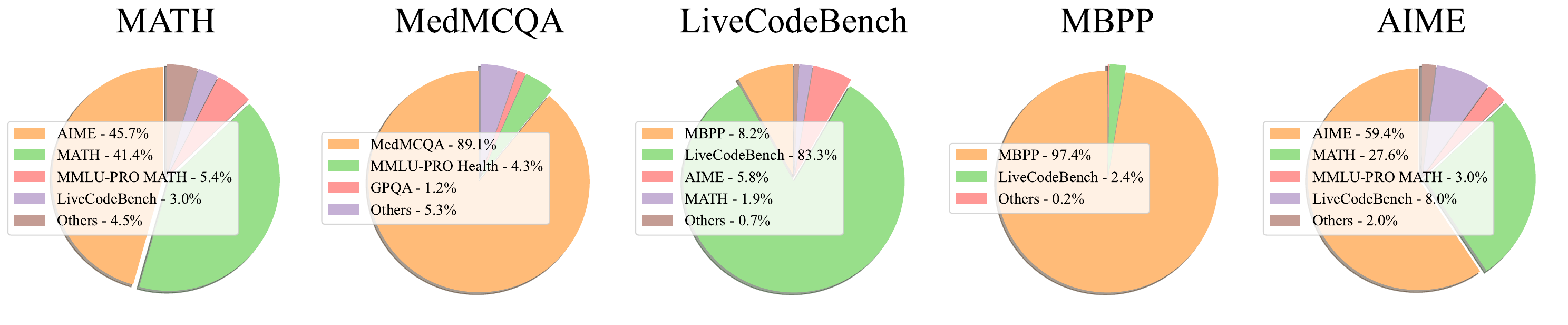}
  \vspace{-4mm}
  \caption{The proportion of support questions retrieved from different source datasets for a given question.}
  \label{fig:prior_ratio}
\end{figure*}

\subsection{Scaling Ability}
\label{Scaling Ability}
To empirically validate the scalability of SMCS framework, we conducted experiments measuring performance improvements with increasing numbers of input LLMs. Fig.~\ref{fig:scale_ability} shows key findings across four standard benchmarks, revealing a clear positive correlation between the scale of input LLMs and overall performance.
For instance, on MMLU-PRO, our SMCS achieves approximately 77\% accuracy with five LLMs. When scaled to ten LLMs, performance improves to nearly 80\%, and with 15 LLMs, the final accuracy approaches 82\%. It not only demonstrates the capability of SMCS to leverage diverse LLMs effectively but also shows that SMCS has the potential to obtain sustainable performance gains as the number of LLMs continually scales up.

\subsection{Out-of-Distribution Performance}
\begin{table}[]
\resizebox{0.47\textwidth}{!}{
\begin{tabular}{l|cccc}
\hline
Question Bank                    & AIME  & MBPP  & GPQA-Diamond & MedMCQA \\ \hline
Single(MMLU-PRO)                 & 73.33 & 82.2  & 64.14        & 75.17   \\
Unified(Eight Datasets)          & 73.33 & 82.80 & 65.15        & 75.75   \\
w/o Question Bank(Random Select) & 56.67 & 82.20 & 56.56        & 74.25   \\ \hline
\end{tabular}}
\caption{Comparison with different question banks.}
\label{tab:ood}
\end{table}
To demonstrate the generalization of the proposed prior selection, we conduct out-of-distribution retrieval experiments. Specifically, we build a question bank using 5,512 questions only from MMLU-PRO and evaluate SMCS on the other four datasets. As shown in Table~\ref{tab:ood}, the results demonstrate that even with a question bank using an out-of-distribution dataset, our retrieval-based prior selection surpasses random selection significantly, while with only marginal performance drops compared with using a multi-dataset question bank. It verifies the strong generalization of our prior selection mechanism when facing out-of-distribution questions. Moreover, SMCS can introduce more diverse questions to further refine LLM capability assessments and boost performance.

\subsection{Analysis on Prior Selection}
\label{Analysis on Retrieval-based Prior Selection}
As shown in Fig.~\ref{fig:prior_ratio}, we display the proportion of support questions retrieved from different source datasets. For a given question, the retrieved support questions are mostly from subjects with similar capability requirements. Specifically, a substantial portion of the support questions are retrieved from the same dataset as the given question, while others are from other datasets with similar subjects. For instance, in the case of MATH, nearly half of the retrieved support questions are from AIME, and for MedMCQA, several are retrieved from the "Health" category in MMLU-PRO. It not only verifies the effectiveness of our proposed method in bridging the given question with relevant prior information but also demonstrates its ability to perform cross-dataset retrieval. This capability significantly increases the amount of relevant prior information, enhancing model assessment and suggesting a potential for scalability. Additionally, we observe a retrieval connection between mathematics and code-generation tasks, e.g., LiveCodeBench and AIME retrieve questions from each other, indicating that solving coding and mathematical problems may require similar capabilities. Moreover, to verify the correlation between prior LLM evaluation and practical performance, we introduce a pairwise ranking score inspired by the ranking loss~\cite{hu2021ranknas,xu2021renas} in Neural Architecture Search (NAS), denoted as 
\begin{equation}
\resizebox{0.8\hsize}{!}{$
\begin{aligned}
    Sco_{rank} = \frac{\displaystyle \sum_{i\in I^{test}}
      \sum_{j\in P}\sum_{k\in N}\mathbf{1}\{V_i^{ref}[j]>V_i^{ref}[k]\}}{\displaystyle \sum_{i\in I^{test}}| \{j|V^{test}_i[j]=0\}|\cdot|\{k|V^{test}[k]=1\}|},
\end{aligned}
$}
\end{equation}
where $I^{test}$ is the index of test questions, $V^{ref}_i$ is the LLM prior vector in Formulation~\ref{formu:prior_vector} for $i_{th}$ test question. $V^{test}_i \in \{0,1\}^{R}$ represents the correctness of all LLMs on the test set, where $R$ is the number of LLMs, $0$ and $1$ indicate an incorrect and correct answer, respectively. As shown in Table~\ref{tab:rank_score}, compared with Symbolic-MoE, our retrieval-based method consistently obtains a higher score, suggesting our method provides a more accurate prior evaluation of LLMs.

\begin{table}[t]
\centering


\resizebox{0.48\textwidth}{!}{
\begin{tabular}{c|ccccc}
\hline
             & MATH  & MBPP  & MedMCQA & LiveCodeBench & AIME  \\ \hline
Symbolic-MoE & 73.54 & 70.2  & 64.62   & 48.64         & 60.64 \\ \hline
SMCS(ours)         & 74.46 & 70.61 & 65.17   & 68.19         & 86.5  \\ \hline
\end{tabular}
}
\vspace{-2mm}
\caption{Pairwise ranking scores of different methods.}
\vspace{-6mm}
\label{tab:rank_score}
\end{table}

\subsection{Analysis on Posterior Enhancement}
\label{Analysis on Posterior Exploration and Exploitation}
\begin{figure}[!t]
  \centering
  \includegraphics[width=\linewidth]{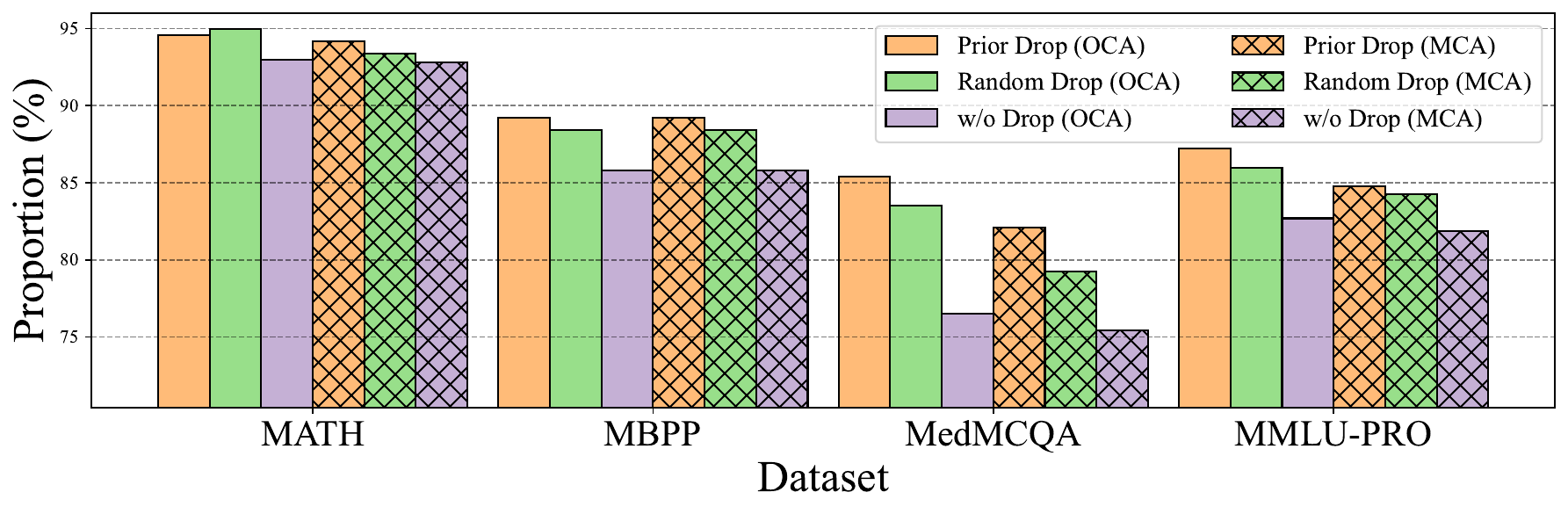}
  \vspace{-8mm}
  \caption{The comparison of different posterior enhancement methods. OCA: One Correct Answer proportion; MCA: Multiple Correct Answers proportion. }
  \label{fig:post_analysis}
  \vspace{-6mm}
\end{figure}

To verify the effectiveness of the proposed posterior exploration and exploitation, we analyze the proportion of correct answers within multiple aggregation responses using different strategies, including the proposed prior drop, random drop, and vanilla aggregating without drop. We use the existing one correct answer proportion (OCA) and the existing multiple correct answers proportion (MCA) to indicate the diversity and quality of multiple aggregating responses, respectively. As shown in Fig.~\ref{fig:post_analysis}, compared with aggregating without drop and random drop, our method can consistently obtain both higher OCA and MCA, suggesting our method can explore a more optimal output region by aggregating multiple times. Thus, there are abundant high-quality candidate responses for exploitation.

\section{Conclusion}
\label{sec:conclusion}
In this paper, to boost the scalability and performance of multi-LLM collaboration systems, we propose SMCS by prior selection and posterior enhancement. Specifically, based on a unified question bank, we propose a retrieval-based prior selection to select the optimal LLMs. Moreover, we propose an exploration–exploitation-driven posterior enhancement, which aggregates references multiple times based on prior information to explore high-quality responses. To select the final output, we propose a hybrid score that combines perplexity and mean pairwise similarity. Extensive experiments demonstrate the effectiveness of SMCS. 

\section*{Limitations}
\label{sec:limitation}
In this section, we discuss the limitations of the proposed SMCS to provide an underlying advance in the field of multi-LLM collaboration systems and to point out promising directions for future research.

\noindent \textbf{Lack of Efficiency Optimization.} To maximize performance upper bounds, SMCS framework does not set constraints on the computational cost of selected LLMs. Thus, the system requires sufficient computational resources and inference time, making it hard to deploy on resource-constrained edge devices. A promising direction for future work is to design multi-LLM systems that optimally balance performance and efficiency.

\noindent \textbf{Lack of Optimization in Inference Configuration.}
In SMCS, all LLMs are queried using the same sampling parameters and prompts. However, a uniform configuration may not be optimal for heterogeneous LLMs within the system. A potential future direction is to tailor prompts and configurations for each LLM individually, which can maximize their capabilities and improve overall system performance.

\section*{Acknowledgements}
This work was supported by the Shanghai Artificial Intelligence Laboratory and a locally commissioned task from the Shanghai Municipal Government.


\bibliography{custom}

\clearpage
\appendix
\section{Appendix}
\subsection{Dataset Details}
\label{app:data_details}
In the experimental section, we evaluate our proposed SMCS framework across eight diverse benchmarks spanning mathematical reasoning, complex question answering, instruction following, and code generation tasks.
Specifically, we construct a balanced test set of 1,196 college-level multidisciplinary questions in MMLU-Pro~\cite{wang2024mmlu} through stratified sampling from the original test set, with 5,512 remaining questions allocated to validation.
For GPQA~\cite{rein2024gpqa}, the diamond subset (graduate-level science questions) serves as our test set, while the remaining data forms the validation set.
For MedMCQA~\cite{pal2022medmcqa}, 1,200 medical professional questions are randomly selected for testing, with 1,000 questions reserved for validation.
MATH-500~\cite{hendrycks2021measuring} subset is used for testing, complemented by 1,000 randomly sampled validation questions from the original dataset.
We employ AIME2024~\cite{AIME2024} as our test set and historical problems (1983-2023) for validation.
For IFEval~\cite{zhou2023instruction} dataset, 300 instruction-following instances are randomly selected for testing, with 241 instances for validation.
The original test set of MBPP~\cite{austin2021program} is preserved for evaluation, while the training and validation sets are combined to form validation.
LiveCodeBench~\cite{jain2024livecodebench} v5 serves as test set, with v6 reserved for validation purposes.

\subsection{LLM Bank Details}
\label{app:llm_bank_details}
To achieve an optimal balance between model diversity and computational efficiency, we carefully curate a collection of 15 mid-sized open-source LLMs (from 20B to 72B) from various architectural families. 
Specifically, the selected LLMs include: Qwen2.5-32B-Instruct~\cite{qwen2.5}, Qwen-2.5-72B-Instruct~\cite{qwen2.5}, Qwen2.5-Coder-32B-Instruct~\cite{hui2024qwen2}, Qwen3-32B~\cite{qwq32b}, GLM-Z1-32B-0414~\cite{glm2024chatglm}, DeepSeek-R1-Distill-Qwen-32B~\cite{deepseekai2025}, DeepSeek-R1-Distill-Llama-70B~\cite{deepseekai2025}, QwQ-32B~\cite{qwq-32b-preview}, Gemma-3-27b-it~\cite{team2024gemma}, TeleChat2-35B-32K~\cite{wang2024telechat}, InternLM2.5-20B-Chat~\cite{cai2024internlm2}, Llama-3.3-70B-Instruct~\cite{grattafiori2024llama}, Llama-3.3-Nemotron-Super-49B-v1~\cite{bercovich2025llama}, HuatuoGPT-o1-72B~\cite{chen2024huatuogpto1medicalcomplexreasoning}, EXAONE-Deep-32B~\cite{exaone-deep}.
As shown in Table~\ref{tab:llm_bank}, our selection encompasses: (1) instruction-tuned variants, and (2) deep thinking models. This strategic composition ensures comprehensive coverage of different capabilities while maintaining manageable computational requirements.

\begin{table}[thbp]

\resizebox{0.5\textwidth}{!}{
\begin{tabular}{l|c|c}
\hline
Name                             & Size & Type              \\ \hline
GLM-Z1-32B-0414~\cite{glm2024chatglm}                 & 32B  & Deep Thinking     \\
Qwen-2.5-72B-Instruct~\cite{qwen2.5}            & 72B  & Instruction-tuned \\
DeepSeek-R1-Distill-Llama-70B~\cite{deepseekai2025}    & 70B  & Deep Thinking     \\
QwQ-32B~\cite{qwq-32b-preview}                       & 32B  & Deep Thinking     \\
Gemma-3-27b-it~\cite{team2024gemma}               & 27B  & Instruction-tuned \\
Qwen2.5-32b-Instruct~\cite{Qwen2.5-32B-Instruct}          & 32B  & Instruction-tuned \\
TeleChat2-35B-32K~\cite{wang2024telechat}                & 35B  & Instruction-tuned \\
Llama-3.3-70B-Instruct~\cite{grattafiori2024llama}          & 70B  & Instruction-tuned \\
EXAONE-Deep-32B~\cite{exaone-deep}                 & 32B  & Deep Thinking     \\
Qwen2.5-Coder-32B-Instruct~\cite{hui2024qwen2}      & 32B  & Instruction-tuned \\
Qwen3-32B~\cite{qwq32b}                       & 32B  & Deep Thinking     \\
Llama-3.3-Nemotron-Super-49B-v1~\cite{grattafiori2024llama} & 49B  & Deep Thinking     \\
DeepSeek-R1-Distill-Qwen-32B~\cite{deepseekai2025}    & 32B  & Deep Thinking     \\
HuatuoGPT-o1-72B~\cite{chen2024huatuogpto1medicalcomplexreasoning}              & 72B  & Deep Thinking     \\
InternLM2.5-20B-Chat~\cite{cai2024internlm2}            & 20B  & Instruction-tuned \\ \hline
\end{tabular}
}
\vspace{-2mm}
\caption{The details of the used LLM bank.}
\vspace{-2mm}
\label{tab:llm_bank}
\end{table}

\subsection{Implementation Details}
\label{app:imp_details}
\noindent \textbf{Inference Configs.} For a fair comparison, we adopt the same inference configs for all experiments. Specifically, we utilize VLLM~\cite{kwon2023efficient} as the framework for LLM inference. For the sampling parameters of LLM inference, we set the temperature to 0.7. The maximum length of output tokens is 8,192 to avoid extremely long responses. We also set the presence penalty to 1.05 to avoid endless repetition. If the length of context tokens exceeds the limitation of an LLM, the YaRN~\cite{peng2023yarn} method is used to extend the context window. Moreover, we use Linq-Embed-Mistral~\cite{kim2024linq} as the embedding model in all experiments, and the embedding dimension is 8,192.

\noindent \textbf{Hyperparameters.} 
For all SMCS experiments, we use nearly the same hyperparameters to ensure consistency and fair comparison. Specifically, we set the number of referencers $K$ as 7. The base retrieval number $N^{sup\_base}$ is 400 while the tolerance threshold coefficient $\gamma = 0.95$. The dropping number $K_{drop}$ is 1. The number of aggregating $n=8$. The balance coefficient of PPL score $\lambda$ is 1.0.   
\begin{figure*}[thbp]
  \centering
  \includegraphics[width=0.9\linewidth]{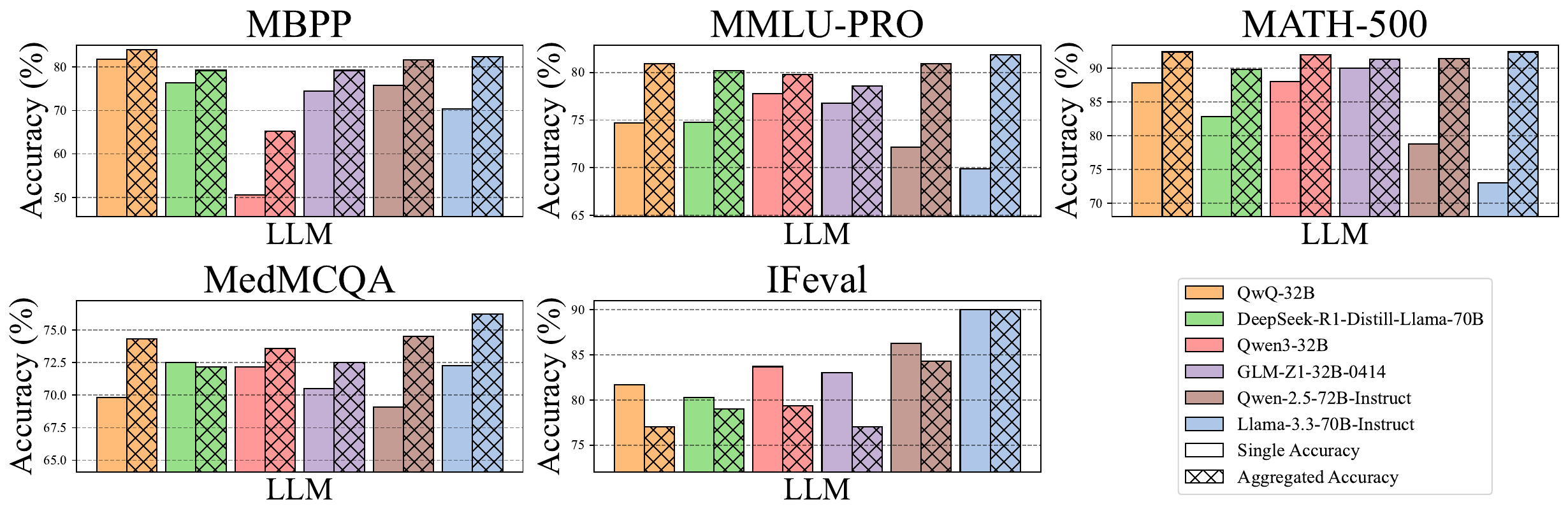}
  \caption{Analysis on aggregator selection with six LLMs across five standard benchmarks}
  \label{fig:agg_select}
\end{figure*}

\noindent \textbf{Compared Methods.} In the experiment, in addition to comparing the performance of single LLMs, we also compared six popular multi-LLMs collaboration methods, and the experimental settings are as follows:
Symbolic\_MOE*~\cite{chen2025symbolic} retains its original model profiling and LLM selection framework while employing Llama-3.3-70B-Instruct for final response aggregation.
MoA~\cite{wang2024mixture} employs 15 LLMs as references, also utilizing Llama-3.3-70B-Instruct as the aggregator.
For both Self-MoA~\cite{li2025rethinking} and Self-Consistency~\cite{chen2023universal}, we utilize each dataset's best LLM to generate 8 responses per query. 
Simple Router directly employs the best-performing LLM from each dataset's question bank for response generation.
Majority Voting~\cite{chen2024more} determines the final output through voting among 15 reference LLMs.

\subsection{Aggregator Selection}
\label{aggregator_selection}
In our SMCS framework, the aggregator plays a pivotal role in consolidating responses from multiple LLMs to generate optimal outputs. To identify the most effective aggregator, we conducted systematic experiments evaluating 6 LLMs as potential aggregators across five diverse benchmarks and the results are shown in Figure~\ref{fig:agg_select}. 
Our analysis revealed that Llama-3.3-70B-Instruct demonstrated consistently superior performance across all datasets, leading to its adoption as our default aggregator.
In addition, two key insights emerged from this experiment: First, we observed a dissociation between single-LLM performance and its aggregation capability on common benchmarks. For instance, while Qwen3-32B outperformed Llama-3.3-70B-Instruct by +8\% on MMLU-PRO, the latter showed significantly better aggregation performance (+4\% over Qwen3-32B).
(2) However, we also identified a positive correlation between single-LLM performance and aggregation capability on the IFEval benchmark. This correlation stems from IFEval's focus on instruction-following tasks, suggesting that optimal aggregator selection should prioritize LLMs with strong instruction-following abilities to maximize MACS performance.

\begin{table*}[thbp]
  \centering
  \vspace{-3mm}
  \resizebox{\textwidth}{!}{
    \begin{tabular}{lccccccccc}
      \toprule
      \textbf{Model} & \textbf{AIME} & \textbf{MATH-500} & \textbf{MBPP} & \textbf{LiveCodeBench} & \textbf{GPQA-Diamond} & \textbf{MMLU-PRO} & \textbf{IFEval} & \textbf{MedMCQA} & \textbf{Avg} \\
      \midrule
      \rowcolor{gray!20} \multicolumn{10}{c}{\textit{Close-source LLMs}} \\
      GPT-o3-mini(2025-01-31)~\cite{gpt-o3-mini} & 73.33 & 84.40 & 62.00 & 54.70 & 66.67 & 74.00 & 82.00 & 74.92 & 71.50 \\
      Claude-3.7-Sonnet(2025-02-19)~\cite{Claude-3.7-sonnet} & 26.70 & 73.20 & 75.40 & 41.30 & 63.64 & 69.43 & 88.00 & 74.75 & 64.05 \\
      GPT-4o(2024-08-06)~\cite{achiam2023gpt} & 10.00 & 74.60 & 74.20 & 29.80 & 52.53 & 73.83 & 82.30 & 76.17 & 59.18 \\
      Claude-3.5-Sonnet(2024-06-20)~\cite{Claude-3.5-sonnet} & 16.70 & 74.20 & 75.80 & 34.30 & 61.62 & 78.34 & 80.30 & 76.00 & 62.16 \\
      GPT-4.1(2025-04-14)~\cite{gpt-4.1} & 50.00 & 85.80 & 79.20 & 42.20 & 67.17 & 80.43 & 86.00 & 80.58 & 71.42 \\
      Close-source Average & 35.34 & 78.44 & 73.32 & 40.46 & 62.33 & 75.21 & 83.72 & 76.48 & 65.66 \\
      \rowcolor{gray!20} \multicolumn{10}{c}{\textit{Open-source LLMs}} \\
      GLM-Z1-32B-0414~\cite{glm2024chatglm}   & 73.33 & 92.2 & 76.20 & 56.50 & 62.12 & 77.84 & 84.30 & 71.08 & 74.20 \\
      Qwen-2.5-72B-Instruct~\cite{qwen2.5} & 16.70 & 80.8 & 77.40 & 26.10 & 46.97 & 72.16 & 86.30 & 69.08 & 59.43 \\
      DeepSeek-R1-Distill-Llama-70B~\cite{deepseekai2025} & 76.70 & 86.00 & 78.40 & 57.80 & 60.60 & 76.09 & 80.70 & 73.17 & 73.68 \\
      QwQ-32B~\cite{qwq-32b-preview} & 80.00 & 91.80 & 85.20 & 59.60 & 62.63 & 78.34 & 82.30 & 70.75 & 76.33 \\
      Gemma-3-27b-it~\cite{team2024gemma} & 30.00 & 84.00 & 70.40 & 27.70 & 50.51 & 65.47 & 81.00 & 64.58 & 59.21 \\
      Qwen2.5-32b-Instruct~\cite{Qwen2.5-32B-Instruct} & 20.00 & 75.60 & 76.00 & 24.00 & 40.91 & 69.15 & 78.70 & 62.92 & 55.91 \\
      TeleChat2-35B-32K~\cite{wang2024telechat} & 10.00 & 70.00 & 70.00 & 19.50 & 33.33 & 67.98 & 82.00 & 57.08 & 51.24 \\
      InternLM2.5-20B-Chat~\cite{cai2024internlm2} & 3.30 & 55.20 & 55.00 & 14.90 & 34.85 & 44.23 & 64.70 & 51.92 & 40.51 \\
      Llama-3.3-70B-Instruct~\cite{grattafiori2024llama} & 30.00 & 73.00 & 70.40 & 30.10 & 46.97 & 69.87 & 90.00 & 72.25 & 60.32 \\
      EXAONE-Deep-32B~\cite{exaone-deep} & 73.33 & 92.20 & 80.60 & 58.10 & 63.13 & 70.48 & 76.30 & 60.83 & 71.87 \\
      Qwen2.5-Coder-32B-Instruct~\cite{hui2024qwen2} & 16.70 & 73.60 & 78.00 & 27.70 & 41.92 & 61.79 & 80.30 & 57.25 & 54.66 \\
      Qwen3-32B~\cite{qwq32b} & 80.00 & 92.80 & 53.20 & 64.10 & 64.65 & 77.76 & 83.00 & 70.92 & 73.30 \\
      Llama-3.3-Nemotron-Super-49B-v1~\cite{grattafiori2024llama} & 16.70 & 75.20 & 65.40 & 28.00 & 48.48 & 67.47 & 82.70 & 70.92 & 56.86 \\
      DeepSeek-R1-Distill-Qwen-32B~\cite{deepseekai2025} & 70.00 & 85.60 & 83.40 & 58.10 & 57.58 & 75.17 & 74.30 & 67.75 & 71.49 \\
      HuatuoGPT-o1-72B~\cite{chen2024huatuogpto1medicalcomplexreasoning} & 13.33 & 73.20 & 78.00 & 24.30 & 52.53 & 74.16 & 74.00 & 76.00 & 58.19 \\
      Open-source Average & 40.67 & 80.11 & 73.09 & 38.37 & 51.45 & 69.86 & 80.04 & 66.18 & 62.47 \\
      \rowcolor{gray!20} \multicolumn{10}{c}{\textit{Ours v.s. Strong Baselines}} \\
      Open-source Upper Bound & 80.00 & 92.80 & 85.20 & 64.10 & 64.65 & 78.34 & 90.00 & 76.00 & 76.33 \\
      \textbf{SMCS(ours)} & \textbf{86.67} & \textbf{94.50} & \textbf{87.00} & \textbf{65.65} & 66.16 & \textbf{81.61} & \textbf{90.00} & 76.50 & \textbf{81.01} \\
      \textit{- v.s. GPT-4.1} & \positive{36.67} & \positive{8.70} & \positive{7.80} & \positive{23.45} & \negative{1.01} & \positive{1.18} & \positive{4.00} & \negative{4.08} & \positive{9.59} \\
      \textit{- v.s. GPT-o3-mini} & \positive{13.34} & \positive{10.10} & \positive{25.00} & \positive{10.95} & \negative{0.51} & \positive{7.61} & \positive{8.00} & \positive{1.58} & \positive{9.51} \\
      \textit{- v.s. GPT-4o} & \positive{76.67} & \positive{19.90} & \positive{12.80} & \positive{35.85} & \positive{13.63} & \positive{7.78} & \positive{7.70} & \positive{0.33} & \positive{21.83} \\
      \textit{- v.s. Claude-3.7-Sonnet} & \positive{59.97} & \positive{21.30} & \positive{11.60} & \positive{24.35} & \positive{2.52} & \positive{12.18} & \positive{2.00} & \positive{1.75} & \positive{16.96} \\
      \bottomrule
    \end{tabular}
  }
    \caption{Main Results on eight mainstream benchmarks using 32,768 maximum output tokens.}  
  \vspace{-3mm}
  \label{tab:main_results_32k}  
\end{table*}
\subsection{Experiments on More Output Tokens}
To further explore the potential of the proposed SMCS with the existing open-source LLMs, we only extend the maximum length of output tokens of referencers and aggregators from 8,192 to 32,768 while retaining the other experiment settings for complex reasoning questions. For a fair and accurate comparison, we also extend the maximum length of output tokens of other LLMs to 32,768. It is worth noting that because non-deep-thinking LLMs respond to questions with fewer output tokens (fewer than 8,192 tokens), we directly utilize the results with 8,192 output tokens for non-deep-thinking LLMs as a comparison. Besides, different from the experimental settings in the manuscript, in coding tasks including MBPP and LiveCodeBench, QwQ-32B is adopted as the aggregator for better performance, while Llama-3.3-70B-Instruct is utilized in other tasks. As shown in Table~\ref{tab:main_results_32k}, with more output tokens, SMCS also maintains remarkable superiority compared with other open-source and closed-source single LLMs. Specifically, based on fifteen mid-sized open-source LLMs, SMCS can surpass the flagship closed-source LLMs GPT-4.1 by 9.59\% and GPT-o3-mini by 9.51\%, respectively. Moreover, under the setting of more output tokens, SMCS can consistently break through the challenging open-source upper bound by 4.68\% and closed-source upper bound by 6.27\%, which demonstrates that SMCS has the potential to push the upper bound of intelligence using multi-LLM collaboration.

\subsection{More OOD Experiments}
To further demonstrate the out-of-domain(OOD) ability of SMCS, we also conduct additional experiments under stricter OOD settings. Specifically, we use the unified question bank constructed from eight datasets as mentioned in Sec.~\ref{sec:exp_setting} and test SMCS in three new OOD datasets, including HumanEval~\cite{chen2021evaluating}, FinQA~\cite{chen2021finqa}, and LiveMathBench~\cite{liu2025your}. As shown in Table~\ref{tab:more_ood}, these new results prove that SMCS maintains strong OOD generalization capabilities as long as it is supported by a question bank with enough diversity.

Most importantly, because our unified question bank was fundamentally designed with scalability, it can seamlessly incorporate an online expansion in real-world deployments. By continuously and dynamically adding new, representative real-world queries to the bank, the system can effortlessly adapt to evolving data distributions and effectively minimize the impact of distributional bias.

\begin{table}[]
\resizebox{0.5\textwidth}{!}{
\begin{tabular}{l|ccc}
\hline
           & HumanEval & FinQA & LiveMathBench \\ \hline
QwQ-32B    & 92.68     & 73.67 & 58.68         \\
GPT-4.1    & 94.51     & 72.97 & 59.50         \\
Self-MoA   & 92.07     & 74.11 & 60.33         \\ \hline
SMCS(ours) & 95.12     & 75.24 & 74.38         \\ \hline
\end{tabular}
}
\caption{Experiments on more OOD settings.}
\label{tab:more_ood}
\end{table}

\subsection{Perplexity Numerical Analysis}
In the Hybrid score of SMCS, the perplexity (PPL) of a response is adopted as the perplexity score, defined as $\mathcal{S}_i^{PPL}=1-PPL(G_i)$. This is then weighted with the similarity score $\mathcal{S}_i^{sim}$ to compute the total score for selecting the final aggregated response. Given that $\mathcal{S}_i^{sim} \in (0,1]$, whereas the theoretical range of $\mathcal{S}_i^{PPL}\in[-\infty,0]$, a potential numerical scale mismatch could arise during the final score calculation.

To investigate whether SMCS suffers from this issue practically, we analyze the statistical distributions of PPL for Llama-3.3-70B-Instruct and QwQ-32B across eight diverse benchmarks. As shown in Table~\ref{tab:stat_ppl}, the empirical PPL values consistently fall within the $[1, 2]$ across all datasets. Consequently, the derived $\mathcal{S}_i^{PPL}$ values strictly reside within $[-1, 0]$, which shares a comparable numerical scale with the similarity score, which means strict normalization is practically unnecessary in most common settings. We attribute this stable behavior to the fact that modern LLMs have undergone extensive pre-training and rigorous post-training, generally exhibiting high confidence in their generated tokens.

\begin{table*}[]
\resizebox{1.0\textwidth}{!}{
\begin{tabular}{l|cccccccc}
\hline
                      & AIME      & MATH-500  & MBPP      & LiveCodeBench & GPQA-Diamond & MMLU-PRO  & IFEval    & MedMCQA   \\ \hline
Llama3.3-70B-Instruct & 1.16±0.06 & 1.07±0.05 & 1.07±0.05 & 1.07±0.05     & 1.12±0.08    & 1.09±0.08 & 1.14±0.12 & 1.13±0.07 \\
QwQ-32B               & 1.44±0.23 & 1.31±0.12 & 1.74±0.19 & 1.86±0.28     & 1.73±0.23    & 1.61±0.23 & 1.87±0.28 & 1.62±0.18 \\ \hline
\end{tabular}
}
\caption{The statistical values of perplexity.}
\label{tab:stat_ppl}
\end{table*}

\subsection{Statistical Analysis}
To provide statements about statistical significance, we conduct repetitive experiments on four datasets, including LiveCodeBench, MMLU-Pro, GPQA-Diamond, and MedMCQA. Each setting is run three times using the hyperparameters in \ref{app:imp_details} under different random seeds. As shown in Table~\ref{tab:statistical_analysis}, SMCS achieves high mean performance across all four datasets, comparable to the results in Table~\ref{tab:main_results}, which demonstrates its ability to deliver consistently strong performance. Furthermore, it can be observed that the standard deviation of SMCS is below 0.6, indicating its superior stability across various settings.

\begin{table}[]
\resizebox{0.5\textwidth}{!}{
\begin{tabular}{l|cccc}
\hline
           & LiveCodeBench & MMLU-PRO   & GPQA-Diamond & MedMCQA    \\ \hline
QwQ-32B    & 39.11±0.63    & 74.69±0.38 & 57.24±0.77   & 69.50±0.33 \\
GPT-4.1    & 42.44±0.37    & 80.41±0.21 & 67.51±0.58   & 80.61±0.21 \\
Self-MoA   & 29.35±0.45    & 69.76±0.40 & 64.65±0.51   & 74.92±0.50 \\ \hline
SMCS(ours) & 52.17±0.46    & 82.05±0.13 & 64.81±0.58   & 75.69±0.42 \\ \hline
\end{tabular}
}
\caption{The statistical analysis on four datasets. Each setting is run three times under different random seeds.}
\label{tab:statistical_analysis}
\end{table}

\subsection{Component Ablation}
\label{component_ablation}
We perform a comprehensive component-wise ablation study on four standard benchmarks to quantify the contribution of each component in our SMCS framework. As shown in Table~\ref{tab:ablation}, the baseline achieves 79.60\% accuracy on MMLU-PRO. Adding the Major Similarity and RPS modules improves performance by +0.67\% and +1.5\%, respectively, reaching 81.43\% when combined. Further gains come from PPL Filtering and Prior Drop, each contributing an additional +0.5\%. Similar improvements are observed on MedMCQA, MATH, and MBPP, confirming the effectiveness of each component in enhancing multi-agent collaboration.

\begin{table}[!t]
\setlength{\belowrulesep}{0pt}
\setlength{\aboverulesep}{0pt}
\centering
\resizebox{0.47\textwidth}{!}{%
\begin{tabular}{cccc|cccc}
\hline
RPS                       & MPS                       & PPL                       & Prior Drop                & MMLU-PRO & MedMCQA & MATH & MBPP \\ \hline
\xmark     & \xmark     & \xmark     & \xmark     & 79.60     & 73.08   & 87.8 & 82.00   \\
\xmark     & \checkmark & \xmark     & \xmark     & 80.27    & 74.00      & 90.00   & 82.2 \\
\checkmark & \xmark     & \xmark     & \xmark     & 81.10     & 75.08   & 91.80 & 82.40 \\
\checkmark & \checkmark & \xmark     & \xmark     & 81.43    & 75.16   & 91.80 & 82.40 \\
\checkmark & \checkmark & \checkmark & \xmark     & 81.52    & 75.42   & 92.40 & 82.60 \\
\checkmark & \checkmark & \checkmark & \checkmark & 82.02    & 75.75   & 92.60 & 82.80 \\ \hline
\end{tabular}
}
\caption{Component ablation on four standard datasets. RPS: Retrieval-based Prior Selection; MPS: Mean Pairwise Similarity; PPL: Perplexity.}
\label{tab:ablation}
\end{table}

\subsection{Performance vs. Cost Study}
Because multi-LLM orchestration lies in maintaining compelling performance gains under realistic computational budgets, we comprehensively evaluate the accuracy vs. compute cost trade-offs by varying three mentioned hyperparameters on the MMLU-PRO dataset: the number of selected referencers($K$), the dropping times ($n$), and the retrieval base number ($N^{sup\_base}$). The results are shown in Tables~\ref{tab:acc_vs_cost_k}, \ref{tab:acc_vs_cost_n} and \ref{tab:acc_vs_cost_N}, which demonstrate that SMCS can obtain a competitive trade-off between effectiveness and efficiency:
\begin{itemize}
\item \textbf{Varying Selected Referencers}($K$ in Table~\ref{tab:acc_vs_cost_k}): As $K$ increases from 1 to 7, accuracy steadily climbs from 79.35\% to a peak of 82.02\%, with a predictable linear increase in cost. Notably, even at a highly constrained budget setting ($K=3$
), SMCS achieves an accuracy of 81.02\% at a cost of only \$0.90. This already significantly outperforms the strong baseline Symbolic-MoE, which achieves 80.60\% accuracy with \$1.71, while reducing the cost by nearly half.
\vspace{-2mm}
\item \textbf{Varying Dropping Times}($n$ in Table~\ref{tab:acc_vs_cost_n}): Increasing the dropping iterations consistently yields performance gains, scaling from 81.27\% ($n=1$) to 82.36\% ($n=64$). The system demonstrates remarkable cost-effectiveness early in the curve and consistently increasing the $n$ will obtain marginal improvement. At $n=8$, SMCS achieves 82.02\% accuracy with a moderate cost of \$1.80, outperforming Self-MoA(69.89\% with \$2.04) and Symbolic-MoE (80.60\% accuracy with \$1.71).
\vspace{-2mm}
\item \textbf{Varying Retrieval Base Number}($N^{sup\_base}$ in Table~\ref{tab:acc_vs_cost_N}): When expanding the retrieval base number $N^{sup\_base}$, the peaking performance is achieved $N^{sup\_base}=200$ with an accuracy of 82.27\%. It can be observed that varying $N^{sup\_base}$ introduces very minimal cost fluctuations. The reason is that a larger $N^{sup\_base}$ may introduce more reasoning LLMs, which will generate more tokens for each response and cause more cost. Extending the base beyond 800 introduces more noise, leading to marginal performance drops, confirming that a moderately sized retrieval base is highly optimal for both cost and accuracy.
\end{itemize}

These exhaustive budget curves clearly illustrate that SMCS does not rely on cost increasing. Under strict, realistic budgets, SMCS comprehensively dominates the strong baselines in both accuracy and cost-efficiency.

\begin{table*}[]
\resizebox{\textwidth}{!}{
\begin{tabular}{l|ccccccccccc}
\hline
K        & 1     & 2     & 3     & 4     & 5     & 6     & 7     & 8     & 9     & 10    & 11    \\ \hline
Acc(\%)  & 79.35 & 79.01 & 81.02 & 81.52 & 81.52 & 81.87 & 82.02 & 81.44 & 81.77 & 81.44 & 81.10 \\ \hline
Cost(\$) & 0.46  & 0.68  & 0.90  & 1.12  & 1.34  & 1.56  & 1.80  & 2.01  & 2.13  & 2.25  & 2.39  \\ \hline
\end{tabular}
}
\caption{The accuracy vs. compute cost curves of selected referencers $K$ on MMLU-PRO.}
\label{tab:acc_vs_cost_k}
\end{table*}

\begin{table}[]
\resizebox{0.5\textwidth}{!}{
\begin{tabular}{l|ccccccc}
\hline
n        & 1     & 2     & 4     & 8     & 16    & 32    & 64    \\ \hline
Acc(\%)  & 81.27 & 81.44 & 81.86 & 82.02 & 82.11 & 82.19 & 82.36 \\ \hline
Cost(\$) & 1.54  & 1.63  & 1.67  & 1.80  & 2.02  & 2.58  & 3.51  \\ \hline
\end{tabular}
}
\caption{The accuracy vs. compute cost curves of dropping times $n$ on MMLU-PRO.}
\label{tab:acc_vs_cost_n}
\end{table}

\begin{table}[]
\resizebox{0.5\textwidth}{!}{
\begin{tabular}{l|cccccccc}
\hline
\textbf{$N^{sup\_base}$} & 10    & 50    & 100   & 200   & 400   & 800   & 1600  & 3200  \\ \hline
Acc(\%)    & 81.44 & 81.86 & 81.94 & 82.27 & 82.02 & 82.11 & 81.27 & 81.19 \\ \hline
Cost(\$)   & 1.73  & 1.73  & 1.76  & 1.79  & 1.80  & 1.86  & 1.90  & 1.97  \\ \hline
\end{tabular}
}
\caption{The accuracy vs. compute cost curves of retrieval base number $N^{sup\_base}$ on MMLU-PRO.}
\label{tab:acc_vs_cost_N}
\end{table}

\subsection{Prompts}
To maximize task-specific performance across diverse benchmarks, we developed customized prompt designs for each of the eight evaluation benchmarks, aligning with their distinctive characteristics, as illustrated in Fig.~\ref{prompt_dataset}. In addition, we elaborated on the prompt design for the aggregator within our SMCS framework by drawing inspiration from the aggregator prompt strategy proposed in MOA~\cite{wang2024mixture}, as shown in Fig.~\ref{prompt_aggregator}.

\clearpage
\newtcolorbox{promptbox1}{
  colback=blue!5!white,
  colframe=blue!30!black,
  fonttitle=\bfseries,
  title=Prompt Design for AIME benchmark,
  boxrule=1pt,
  arc=3pt,
  boxsep=5pt,
  left=6pt,  
  enhanced jigsaw  
}

\newtcolorbox{promptbox2}{
  colback=blue!5!white,
  colframe=blue!30!black,
  fonttitle=\bfseries,
  title=Prompt Design for MATH benchmark,
  boxrule=1pt,
  arc=3pt,
  boxsep=5pt,
  left=6pt,  
  enhanced jigsaw  
}

\newtcolorbox{promptbox3}{
  colback=blue!5!white,
  colframe=blue!30!black,
  fonttitle=\bfseries,
  title=Prompt Design for MBPP benchmark,
  boxrule=1pt,
  arc=3pt,
  boxsep=5pt,
  left=6pt,  
  enhanced jigsaw  
}

\newtcolorbox{promptbox4}{
  colback=blue!5!white,
  colframe=blue!30!black,
  fonttitle=\bfseries,
  title=Prompt Design for LiveCodeBench benchmark,
  boxrule=1pt,
  arc=3pt,
  boxsep=5pt,
  left=6pt,  
  enhanced jigsaw  
}

\newtcolorbox{promptbox5}{
  colback=blue!5!white,
  colframe=blue!30!black,
  fonttitle=\bfseries,
  title=Prompt Design for GPQA benchmark,
  boxrule=1pt,
  arc=3pt,
  boxsep=5pt,
  left=6pt,  
  enhanced jigsaw  
}

\newtcolorbox{promptbox6}{
  colback=blue!5!white,
  colframe=blue!30!black,
  fonttitle=\bfseries,
  title=Prompt Design for MMLU-PRO benchmark,
  boxrule=1pt,
  arc=3pt,
  boxsep=5pt,
  left=6pt,  
  enhanced jigsaw  
}

\newtcolorbox{promptbox7}{
  colback=blue!5!white,
  colframe=blue!30!black,
  fonttitle=\bfseries,
  title=Prompt Design for IFEval benchmark,
  boxrule=1pt,
  arc=3pt,
  boxsep=5pt,
  left=6pt,  
  enhanced jigsaw  
}

\newtcolorbox{promptbox8}{
  colback=blue!5!white,
  colframe=blue!30!black,
  fonttitle=\bfseries,
  title=Prompt Design for MedMCQA benchmark,
  boxrule=1pt,
  arc=3pt,
  boxsep=5pt,
  left=6pt,  
  enhanced jigsaw  
}

\newcommand{\ccite}[2][red]{\textcolor{#1}{\cite{#2}}}
\newtcolorbox{promptbox9}{
  colback=blue!5!white,
  colframe=blue!30!black,
  fonttitle=\bfseries,
  title=Prompt Design for Aggregator,
  boxrule=1pt,
  arc=3pt,
  boxsep=5pt,
  left=6pt,  
  enhanced jigsaw  
}
\begin{figure*}[hbp]  
\centering
\begin{promptbox1}
\textbf{System Prompt:} "Please reason step by step, and put your final answer within \textbackslash\textbackslash boxed\{\}." \\
\textbf{User Prompt:} "Question: \{question\}."
\end{promptbox1}

\begin{promptbox2}
\textbf{System Prompt:} "You are a math problem solver. Please solve the following math problem. Be sure to explain your solution in detail. The numerical values in the answer should be surrounded by \textbackslash\textbackslash boxed{}. The final answer should start with 'The answer is' and give the conclusion directly. Do not add any extra content." \\
\textbf{User Prompt:} "Question: \{question\}."
\end{promptbox2}

\begin{promptbox3}
\textbf{System Prompt:} "You are an exceptionally intelligent coding assistant that consistently delivers accurate and reliable responses to user instructions." \\
\textbf{User Prompt:} "Question: \{question\}."
\end{promptbox3}

\begin{promptbox4}
\textbf{System Prompt:} "You are an expert Python programmer. You will be given a question (problem specification) and will generate a correct Python program that matches the specification and passes all tests." \\
\textbf{User Prompt:} "Question: \{question\}."
\end{promptbox4}

\begin{promptbox5}
\textbf{System Prompt:} "You are a very intelligent assistant, who follows instructions directly." \\
\textbf{User Prompt:} "Question: \{question\}."
\end{promptbox5}

\begin{promptbox6}
\textbf{System Prompt:} "The following are multiple choice questions (with answers) about {}. Think step by step and then output the answer in the format of "The answer is (X)" at the end." \\
\textbf{User Prompt:} "Question: \{question\}."
\end{promptbox6}

\begin{promptbox7}
\textbf{User Prompt:} "Instruction: \{question\}."
\end{promptbox7}

\begin{promptbox8}
\textbf{System Prompt:} "Provide your step-by-step reasoning first, and then print "The answer is (X)", where X is the answer choice (one capital letter), at the end of your response." \\
\textbf{User Prompt:} "Question: \{question\}."
\end{promptbox8}
\vspace{-2mm}
\caption{Prompt Design for eight diverse benchmarks within our SMCS framework.}
\label{prompt_dataset}
\end{figure*}

\begin{figure*}[hbp]  
\centering
\begin{promptbox9}
\textbf{System Prompt:} "You have been provided with a set of responses from various open-source models to the latest user query. Your task is to synthesize these responses into a single, high-quality response. It is crucial to critically evaluate the information provided in these responses, recognizing that some of it may be biased or incorrect. Your response should not simply replicate the given answers but should offer a refined, accurate, and comprehensive reply to the instruction. Ensure your response is well-structured, coherent, and adheres to the highest standards of accuracy and reliability.  \\
Responses from models: \\
1.\{Response1\} \\
2.\{Response2\} \\
...
" \\
\textbf{User Prompt:} "Question: \{question\}."
\end{promptbox9}

\vspace{-2mm}
\caption{Prompt Design for Aggregator within our SMCS, inspired by MoA~\cite{wang2024mixture}.}
\label{prompt_aggregator}
\end{figure*}

\end{document}